# Tracing liquid level and material boundaries in transparent vessels using the graph cut computer vision approach

*Sagi Eppel[1]*

## Abstract

Detection of boundaries of materials stored in transparent vessels is essential for identifying properties such as liquid level and phase boundaries, which are vital for controlling numerous processes in the industry and chemistry laboratory. This work presents a computer vision method for identifying the boundary of materials in transparent vessels using the graph-cut algorithm. The method receives an image of a transparent vessel containing a material and the contour of the vessel in the image. The boundary of the material in the vessel is found by the graph cut method. In general the method uses the vessel region of the image to create a graph, where pixels are vertices, and the cost of an edge between two pixels is inversely correlated with their intensity difference. The bottom 10% of the vessel region in the image is assumed to correspond to the material phase and defined as the graph and source. The top 10% of the pixels in the vessels are assumed to correspond to the air phase and defined as the graph sink. The minimal cut that splits the resulting graph between the source and sink (hence, material and air) is traced using the max-flow/min-cut approach. This cut corresponds to the boundary of the material in the image. The method gave high accuracy in boundary recognition for a wide range of liquid, solid, granular and powder materials in various glass vessels from everyday life and the chemistry laboratory, such as bottles, jars, Glasses, Chromotography colums and separatory funnels.

## 1. Introduction

Many types of material such as liquids, powders and granules are dealt with almost exclusively while carried inside transparent vessels (bottles/jars) or on top of carrier vessels (spatula/plates). Dealing with such materials demands the ability to accurately identify their location and boundaries within the vessel. Visual recognition of material interfaces is essential for determining properties such as liquid level and volume as well as the recognition of processes such as phase separation, precipitation and evaporation. Applications for methods that can automatically find such boundaries range from industrial bottle filling to everyday life beverage handling. One of the fields in which such recognition is particularly important is chemistry laboratory, where interface recognition is essential of controlling numerous laboratory processes such as extraction, distillation, crystallization and column chromatography.[1] Automatic recognition of phase boundaries is therefore essential for automation of large segments of chemistry research .[2-14]

[1] Employee of Cortica Israel, sagieppel@gmail.com

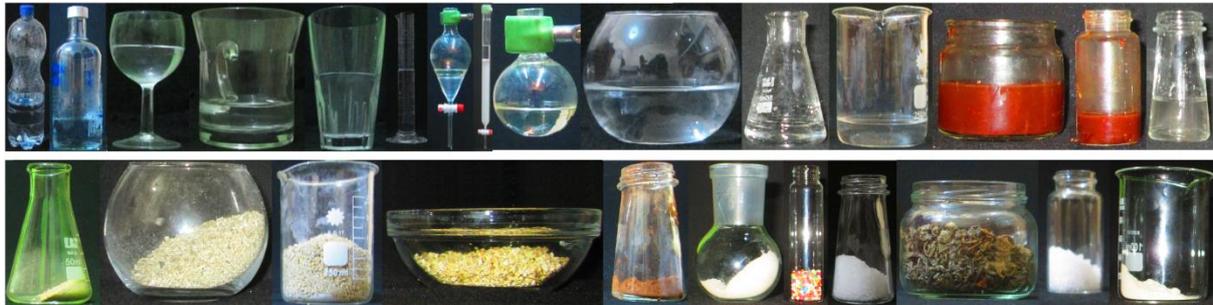

**Figure 1. Example of various fluids (above) and solids (below) in transparent containers taken from everyday life and from chemistry laboratories**

Several automatic approaches have been suggested so far for tracing the boundary of materials in transparent vessels [5, 7, 13-44] and are discussed in Section 1.1. However, these approaches are mostly limited to tracing a single straight line corresponding to liquid level or are based on unique algorithms that make them slower and harder to implement. A general approach that can on one hand be applied to various materials with unrestricted surface shapes and on the other have efficient and fast implementation is still missing. This work presents a new computer vision method for tracing the boundaries of materials in transparent vessels using the graph cut algorithm.[45-49] The method receives an image of a transparent vessel containing some materials and the boundaries of the vessel in the image. It then traces the boundary of the material in the vessel using the graph cut method. The only assumption is that the first phase (hence the material) covers the bottom 10% of the vessel but not the top 10% of the vessel region in the image. Hence, the boundary of the material passes above the bottom 10% pixels in the vessel region of the image and below the top 10% pixels in the vessel region of the image. Finding the material boundary curve was achieved using the max-flow/min-cut approach[46] that transforms the image into a graph and splits it between two regions one is defined as the graph source and the other as the graph sink. The first step involves defining the bottom 10% pixels of the vessel as corresponding to the graph source and the top 10% pixels of the vessel as corresponding to the graph sink.[46, 49] In the second step the min-cut/max-flow approach is used to find the best curve along the vessel region of the image that splits the graph between the sink and source region, this curve is then defined as the material boundary in the image.[45, 46, 49] This method was examined on images of various materials and vessels. The results show a fast and high accuracy recognition of phase boundaries for various cases. However, the lack of physical constraint on the boundary shape and the assumption that the materials completely cover the vessel bottom are the two main sources of errors.

## 1.1. Previous approaches for fill level and phase boundary determination

Various approaches for recognitions of material boundaries have been explored so far, mostly for application of liquid level recognition in industrial bottle filling. These approaches include the use of capacitors or laser beams which identify the changes in the dielectric or reflectance in the liquid-air interface. Another set of approaches uses machine vision, which demands nothing more than a camera. A computer vision-based approach for boundary recognition is usually based on identifying edges or lines of the strong intensity gradient in the image. [27-30, 32-35] An alternative to this approach is the use of assisted computer vision that uses colored floating beads[5, 7, 16] or structured light[17-19] for tracing the liquid interface. These approaches are mostly limited to recognition of a single line and hence are restricted to recognition of liquid level or the phase boundaries of viscous fluid viewed from flat angles. Slightly more sophisticated methods are based on scanning the image of the vessel line by line and finding the parabolic curve which best correlates with the liquid interface.[13] This approach allows recognition of the liquid surface from various angles, but is still limited to liquids with flat surfaces. Recently the Desikjara algorithm was applied for the recognition of material boundaries with no flat surfaces.[14] This method scans for an optimal curve between two pixels on the vessel contour in the image and define this curve as the phase boundary.

## 1.2. The advantage and disadvantage of the graph approach in restricted segmentation of materials in transparent containers

Applying the graph cut approach to the problem of boundary recognition has several major advantages, which include:

a) The ability to trace the boundaries of unrestricted boundary shapes, which makes it effective for materials with unrestricted surface shapes such as solids and powders.

b) The ability to find the globally optimal solution (similar to Desikjara[14]) in nearly real time.[50]

c) Strong theoretical background, and freely available code.[46]

One limitation of the methods is the need to predefine image regions of the image corresponding to each phase (material and air) before segmentation. Another limitation is the relative difficulty in defining physical constraints to the boundary shape.

## 2. The graph cut method

The graph cut approach has emerged as one of the most efficient methods of tracing the boundaries of objects in images. This approach has been discussed in a large number of papers[45-49, 51] and will be summarized here briefly. The graph cut method is derived from graph theory, as a set of methods for splitting a single connected graph into two or more disjoint graphs with a minimum separation cost.[45, 46] In general, we defined a graph $G(E,V)$ as a set of vertices ($V$, Figure 2a) connected by edges ($E$). Each edge connects two vertices and has a specific cost (or weight). Two graphs are defined as disjointed if there is no edge that connects any of the vertices in one graph to any vertex in the other graph, and there is no path along a set of edges that can lead from any vertex in one graph two to a vertex in the second graph (Figure 2b).

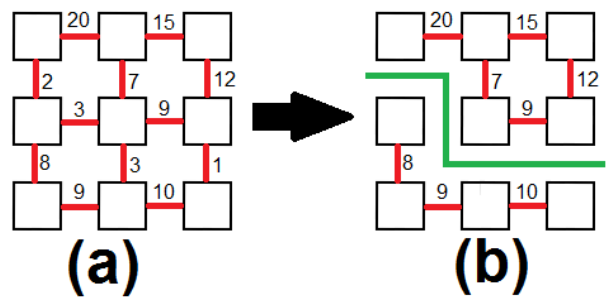

**Figure 2. a) A graph is defined as a set of vertices (squares) connected by edges with specific costs/weights (red lines). b) The graph cut approach finds the best way to split the graph into two disjointed graphs by removing edges with a minimal cost.**

Separating/cutting one connected graph into two disjointed graphs is done by removing all edges that link the two graphs (Figure 2). The graph-cut method involves finding the cut with the smallest cost that separates one graph into two disconnected graphs (Figure 2b). The cut cost is simply the sum of the costs of all the edges that were removed to create this cut.

### 2.1. Graph cuts for image segmentation

The graph cut approach can be used in image segmentation by using the image as a graph where the pixels correspond to vertices (Figure 3). Edges correspond to the similarity between a neighboring pair of pixels (Figure 3b), and their costs are proportional to the similarity in between these two pixels. The general idea is that the edges cost should be high between similar regions (or pixels) corresponding to the same object and low between dissimilar regions corresponding to different objects. As a result, min-cut will represent the best segmentation of the image between different objects or materials (Figure 3c-d). Material boundaries in an image are mostly characterized by a sharp change in colour or intensity. Therefore, the cost of an edge between two adjacent pixels was defined as inversely related to their intensity difference, which encourage splits between dissimilar regions.

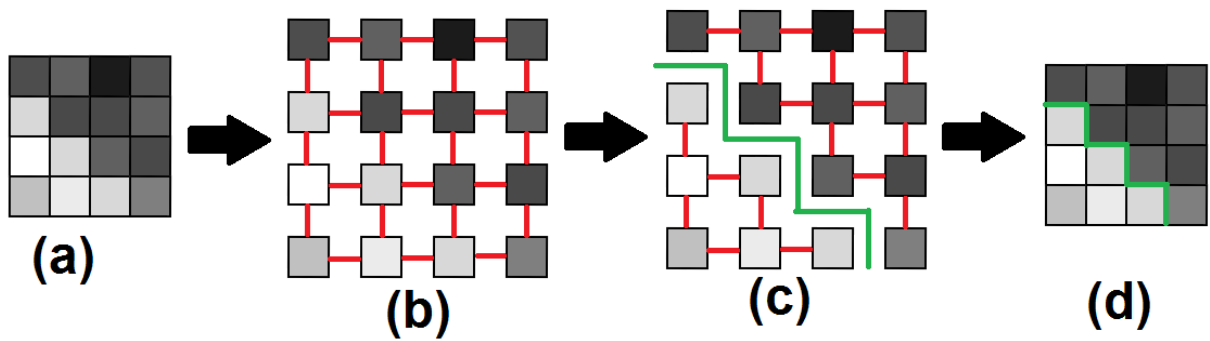

**Figure 3. Graph cut approach for image segmentation. a-b) Image is defined as a graph with pixels as vertices and edges between neighboring pixels. The cost of an edge between a pair of pixels is inversely related to the intensity difference between these pixels. c) The graph is split using the graph cut approach. d) The best cut is used as the boundary in the image.**

## 2.2. The max-flow/min-cut method

One of the most common variations of the graph cut is the max-flow/min-cut approach.[45, 46, 48] In this mode the graph contains two additional vertices that are referred to as source($s$) and sink($t$) (Figure 4a). The min-cut finds the minimal cut that splits the graph between the sink and source (Figure 4b).

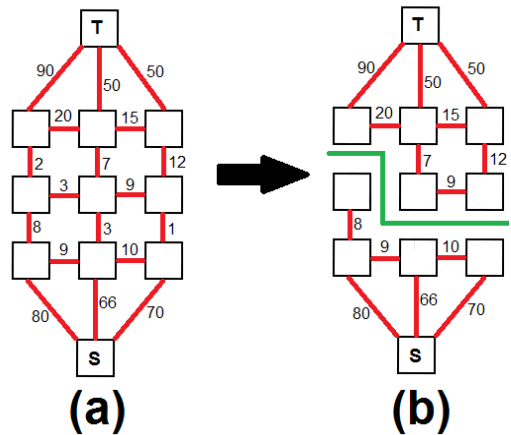

**Figure 4.** The max-flow/min-cut method: two additional vertices of source(*s*) and sink(*t*) are added and the cheapest cut that splits the graph between these two vertices is found.

## 2.3. Max-flow/Minimal-cut approach for image segmentation

The max-flow/min-cut method can be applied for segmenting an image into two distinct regions, such as object and background or material and air. This is done by defining one region of the image as corresponding to the sink vertex and another as corresponding to the source vertex of the graph and using the graph cut to split the image between these regions (Figure 5).[45, 46, 51]

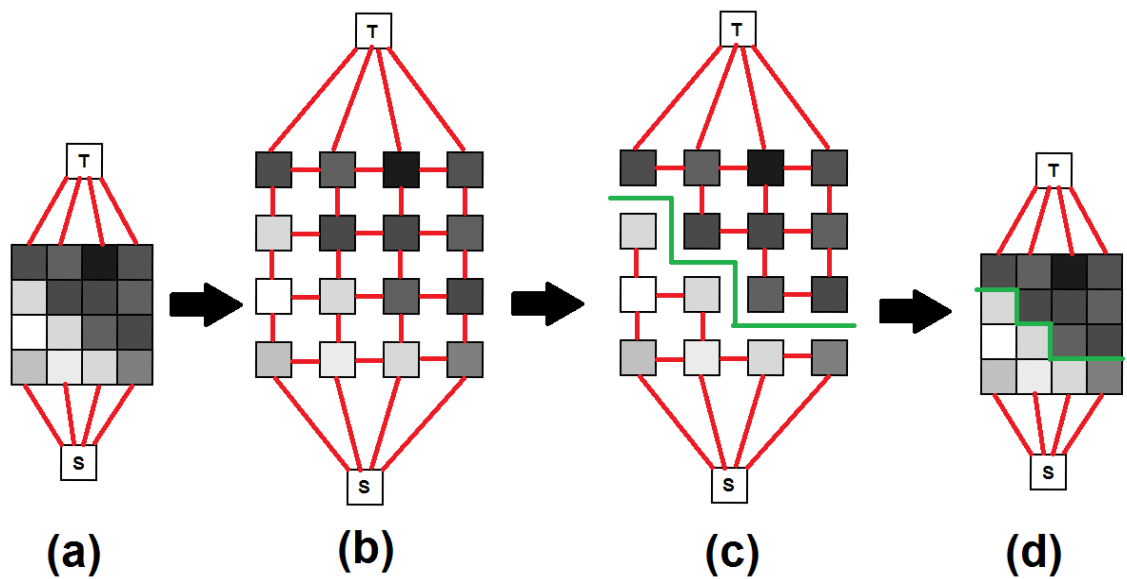

**Figure 5.** Using the max-flow/min-cut approach for image segmentation. a) Define set of pixels in the image belonging to one phase as connected to the source vertex, and set of pixels belonging to a second phase as connected to the sink vertex. b) Define a graph with pixels as vertices. c) Use the max-flow/min-cut approach to split the graph between the sink and source. d) Use the cut as the boundary between the two regions.

For example, in many cases in which materials are handled in transparent vessels, it is possible to assume that the bottom 10% of the vessel is completely covered by the material (liquid or solid) while the top fraction of the vessel in the image is empty (air). By assuminmg that the pixels in the vessel bottom are connected by edges with infinite cost to the source vertex (Figure 5a) and the pixels in the top region of the vessel are connected by edges with infinite cost to the sink vertex (Figure 5a). The min-cut/max-flow method can be used to find the best cut that separates the source and the sink region, and use the cut as the boundary of the material in the vessel. The main limitation of this method is that it's necessary to predefine an image region corresponding to each phase beforehand.[45, 46, 51]

## 2.4. Applying the graph cut approach for finding material boundaries in transparent vessels

The max-flow/min-cut method was applied to tracing the boundary of materials in transparent vessels by applying the following four steps (Figure 6):

a) Receive input of an image of a transparent vessel containing a material and the boundary of the vessel in the image (Figure 6a) and define a graph composed of all the pixels inside the vessel region of the image as vertices (Figure 6b).

b) The cost of all edges between pixels was defined as zero for nonadjacent pixels and inversely related to their intensity difference for adjacent pixels. The exact cost function is discussed in Section 3. The cost of each edge in the graph was divided by the width of the vessel in the edge row. This was done in order to prevent favoring cuts along with a narrow region of the vessels. In addition, the costs of all horizontal edges were increased by a factor of 1.3 to discourage vertical cuts.

c) Two additional vertices of source and sink were added to the graph. The source vertex was defined as related to the material phase while the sink vertex was defined as related to the air phase. The bottom 10% pixels in the vessel region of the image were connected to the source vertex by edges of infinite cost. The top 10% pixels in the vessel region were connected to the sink vertex by edges with infinite cost (Figure 6c).

d) The max-flow/min-cut is used to find a cut with minimal cost that separates the graph between the sink and source vertices (Figure 6d). These cuts represent the boundary of the material in the image (Figure 6d).

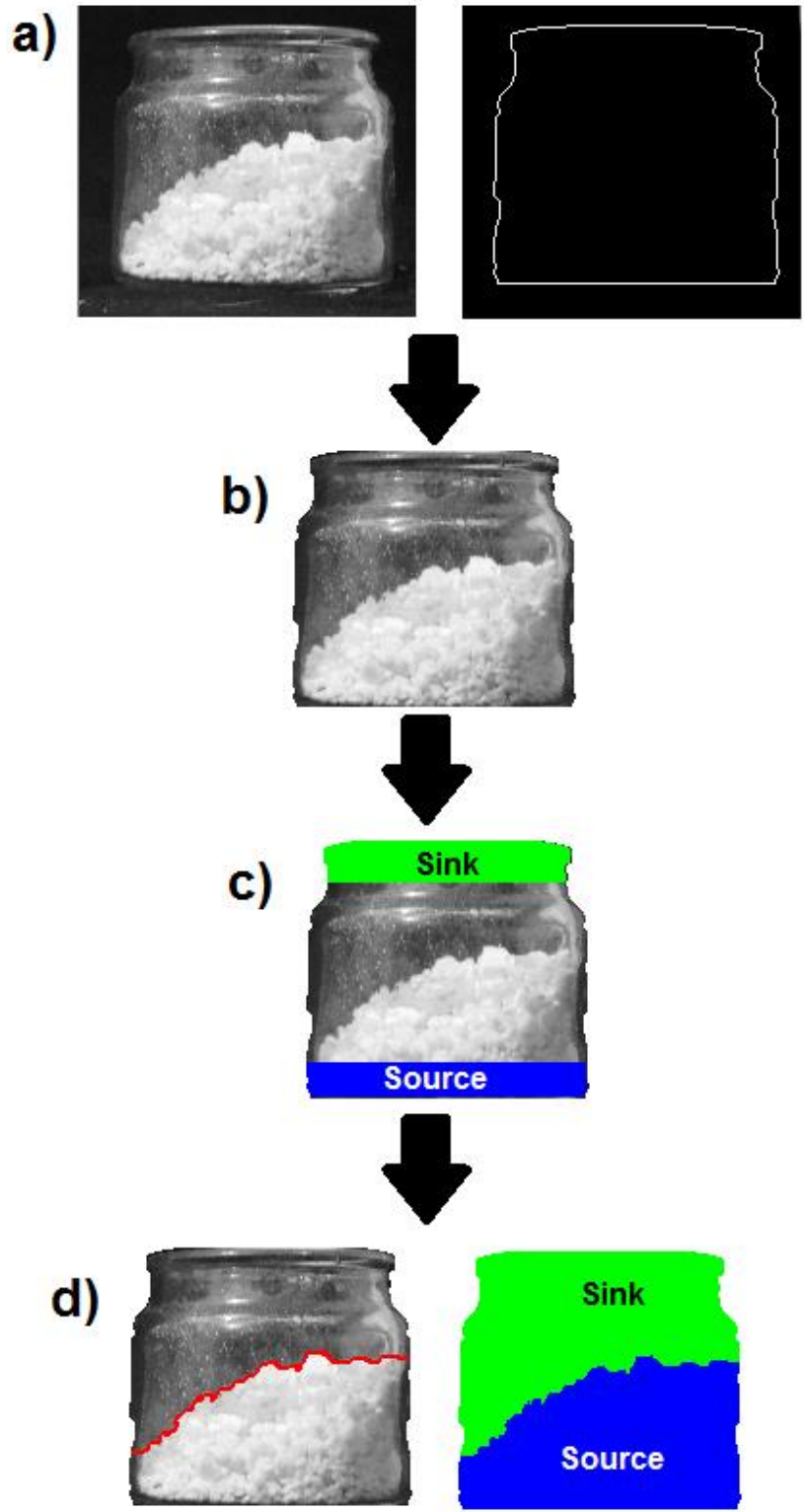

**Figure 6. Scheme of the method for tracing boundary of materials in transparent vessels using graph cut. a) Receive an image of a transparent vessel containing material and the contour of the vessel in the image. b) Create a graph using all the pixels in the vessel region of the image as vertices. c) Define the top pixels/vertices in the vessel region of the image as the graph sink and the bottom pixels as the graph source. d) Use the max-flow approach to find the best split between the graph sink and source. This split is the boundary of the material in the image.**

## 3. Cost function for edges

High intensity or colour difference between image regions represent a strong indicator for the existence of boundaries between two materials and objects. Cuts between regions with large intensity differences are therefore likely to represent a material boundary in the image. In order to encourage cuts between regions with a large difference in intensity, the cost of an edge between two neighbouring pixels was defined as inversely correlated to their intensity difference. The simplest form of such relation is:

1. Cost(i,j)=-|I(i)-I(j)|.

Where Cost(i,j) is the cost of the edge between adjacent pixels *i* and *j*. *I(i)-I(j)* are the intensities of pixels *i* and *j* respectively. A more robust cost function is the exponential function is the exponential function:

2. $\text{Cost}(i,j) = e^{-(\frac{I(i)-I(j)}{2\sigma})^2}$

Where σ is the standard deviation of intensity in the image and can be used as an adjustment parameter. This function is more widely used due to its robustness and the fact that it is more representative for color distribution in the real world.[45, 46, 49] In order to accelerate computation and increase simplicity, edges were set only between each pixel and its four direct neighbors. A few more additions were made to the edge cost and described below.

**a) Width normalization:** Cutting the graph along narrow regions of the vessel involves removing much fewer edges than cuts along wide regions of the vessel. As a result the cut cost in a given region of the image is directly related to the vessel width in this region which can lead to bias toward cuts in cuts in narrow vessel regions. To solve this, the cost of each edge was divided by the width of the vessel in the row of this edge.

**b) Increasing cost of horizontal edges:** Materials that are handled in transparent containers consist mostly of fluids, powders or small particles. Under normal conditions, these material surfaces tend to be flat or of limited steepness. Increasing the cost of horizontal edges by some factors discourages vertical cuts and as a direct result discourages boundary curves of high steepness.

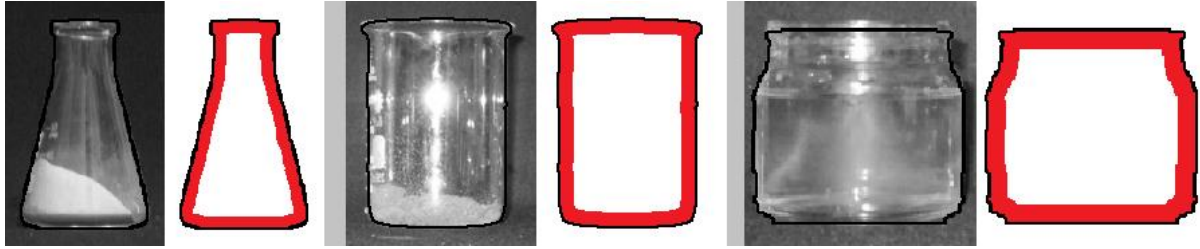

**Figure 7. Penalty zone: Image areas near the vessel boundary (marked black) are more likely to cause false recognition, therefore, all pixels within a certain distance of the vessel boundary (red) are considered located in a penalty zone where the cost of the edges is tripled.**

**c) Penalty zone:** In addition to the above assumption, it is often desirable to discourage cuts along specific regions of the vessel with high visual disturbance.[14, 43] For example, pixels with a close vicinity to the boundary of the vessel tend to have strong intensity gradients due to the high slope of the vessel surface in this region. This strong gradient can cause this region to be falsely identified as the phase boundary (Figure 7). To discourage cuts along these regions, the cost of all edges in these regions was tripled (Figure 7).

**Table 1: Detection rate of material boundaries in images of vessels containing solids and liquids for various of edges cost functions (Section 3).**

|  | Detection | |
| --- | --- | --- |
| $\sigma$ [1] | Liquids | Solids |
| 10 | 73% | 55% |
| 20 | 83% | 62% |
| 30 | 81% | 58% |
| 40 | 81% | 53% |
| 50 | 81% | 45% |
| 60 | 80% | 35% |
| 70 | 76% | 33% |
| 80 | 72% | 28% |
| 90 | 70% | 24% |
| 100 | 66% | 23% |
| Linear[2] | 76% | 18% |

[1] **$\sigma$ Used in the exponential cost function (Section 3). The pixels intensities are in value range of 0-255.**

[2] **Linear cost function (Section 3).**

## 4. Evaluation method

The recognition method was tested using a set of 251 images containing transparent vessels with various materials. This set contained 150 images of vessels containing solid materials and 101 images of vessels containing fluids. The glass containers used in the image included ordinary glass vessels (i.e., jars, bottles and cups) as well as glassware used for analytical chemistry and organic synthesis (i.e., beakers, chromatographic columns, separatory funnels, Erlenmeyer flasks, round-bottom flasks, and vials).[1] The solids in the vessels included various powders, grained materials, and dry leaves with various particle sizes and morphologies; certain solids were immersed in liquids to examine the recognition of liquid-solid interfaces. The liquids used in the images included water, oil, silica slurries, and various organic solvents (DMF, hexane, etc.). All pictures were taken using a uniform, black, and smooth curtain fabric with no folds as a table cloth and background. The areas belonging to the vessel in the image were automatically recognized using template matching or by extracting the vessel region from the uniform background based on its symmetry (See supporting material). The C++ source codes for the program are supplied in the supporting material. The graph cut was applied using the Boykov-Kolmogorov algorithm[46] supplied freely (See supporting material). The image sets used for the testing of the method are supplied in the supporting materials.

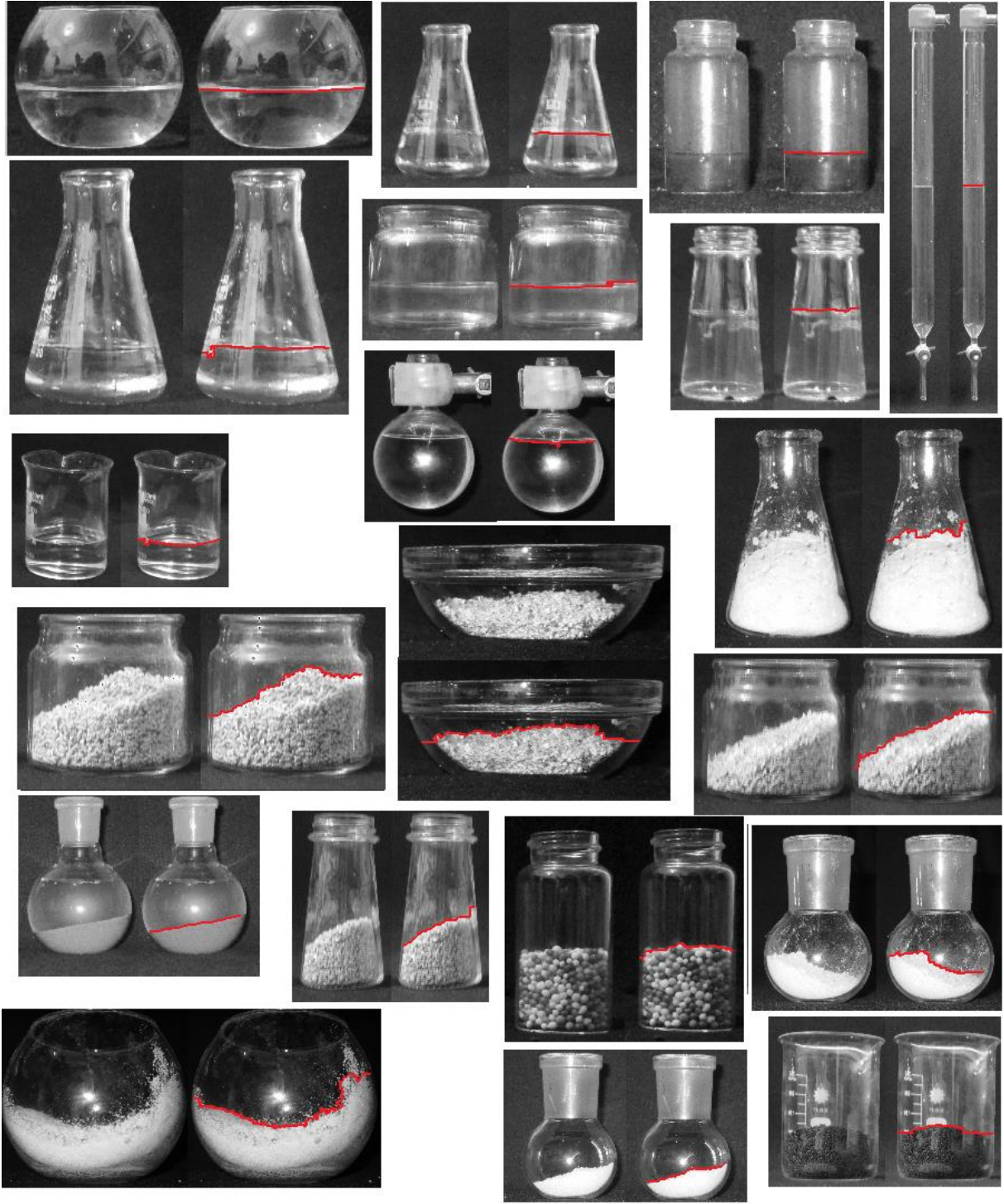

**Figure 8. Examples of good detections. Detected phase boundaries marked red. These results are for a system with exponential cost function and sigma=20.**

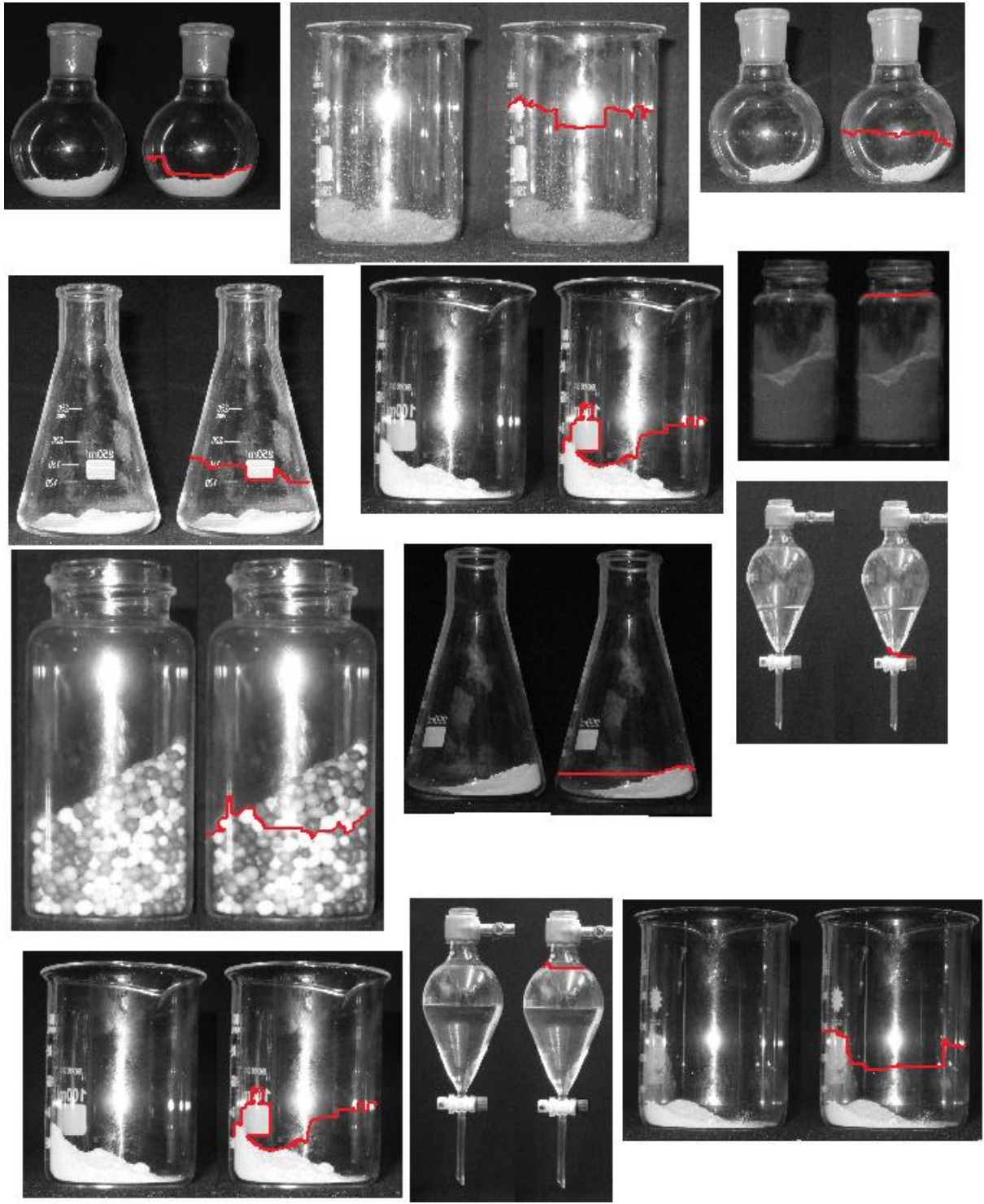

**Figure 9. Examples of misdetections. Misdetected phase boundaries are marked red. These results are for a system with exponential cost function and sigma=20.**

# 5. Results and discussion

The results of the method in boundary recognition are shown in Figures 8-9 and Table 1. The method gave high detection for liquids and solids materials in various vessels and illuminations (Figure 8). It can be seen from Table 1 that the exponential method for edge cost evaluation (Section 3) gave superior results for solid material boundary recognition while for liquids both linear and exponential edge cost functions (Section 3) scored the same. A main source of misdetections (Figure 9) are cases in which the vessel contains a small amount of materials which cover less than 10% of the vessel bottom. In this case the main assumption of the method that the bottom fraction of the vessel in the image corresponds to the material phase is invalid and the result is misdetection. Another main limitation of the method is the lack of physical constraint on the shape of the cut, which can lead to physically unlikely phase boundaries (Figure 9). Another source of misdetection is surface reflections, and functional parts of the vessel which involve strong edges that are mistakenly identified as the boundary of the material in the vessel (Figure 9). Yet another main source of misdetection are materials with strong texture which can lead to strong edges within the material bulk (Figure 9). This texture is often mistaken for the material boundary. Comparing these results to those Desirja algorithms (presented in previous work) for the same set of images shows that the accuracy of the graph cut method is identical to the Desirja[43] for liquid materials (83% vs 82%) but much lower for solids (62% vs 88%). This difference probably stems from cases with materials that cover less than 10% of the vessel bottom as well as the lack of physical constraint on the material boundary.

# 6. Conclusion

The graph cut approach gave high recognition accuracy for tracing the materials boundaries in transparent vessels. The assumption that the bottom of the vessel is completely covered by the material while the vessel top is empty is the main limitation of this method and fails for vessels containing small quantities of materials. The lack of physical constraints represents the second limitation. The running time of the method is near real time, making this approach very useful for real time tracking of material boundaries. Such a method could be useful in areas such as chemistry laboratory automation, and any field in which materials are handled in transparent vessels.

# 7. Supporting material

C++ source code for the method used here is available at: https://github.com/sagieppel/Tracing-liquid-level-and-material-boundaries-in-transparent-vessels-using-the-graph-cut--maxflow-mod

C++ source code for the method described in this work is supplied as supporting material. This code use the Boykov-Kolmogorov algorithm[46] which can be downloaded from http://vision.csd.uwo.ca/code/

The method demand as input image of the vessel as well as the the boundary of the vessel in the image as a binary edge file with the vessel boundaries marked as 1. Source code (Matlab) for automatic tracing the vessel boundary in the image using segmentation from background or template available at:

1) www.mathworks.com/matlabcentral/fileexchange/46887-find-boundary-of-symmetricobject-in-image

2) www.mathworks.com/matlabcentral/fileexchange/46907-find-object-in-image-usingtemplate--variable-image-to-template-size-ratio-

Image sets used for the testing of the code are available at: Larger image sets available at:

https://goo.gl/photos/JzNJHejDJXh4bPub8

https://goo.gl/photos/V1nMfiox2L5GuJY36

https://flic.kr/s/aHsktsKrfs

https://flic.kr/s/aHsksFjwjn